# Multi-modal Predictive Models of Diabetes Progression

Ramin Ramazi
University of Delaware
ramazi@udel.edu

Christine Perndorfer
University of Delaware
cperndorfer@psych.udel.edu

Emily Soriano
University of Delaware
esoriano@psych.udel.edu

Jean-Philippe Laurenceau
University of Delaware
jlaurenceau@psych.udel.edu

Rahmatollah Beheshti
University of Delaware
rbi@udel.edu

## ABSTRACT

With the increasing availability of wearable devices, continuous monitoring of individuals' physiological and behavioral patterns has become significantly more accessible. Access to these continuous patterns about individuals' statuses offers an unprecedented opportunity for studying complex diseases and health conditions such as type 2 diabetes (T2D). T2D is a widely common chronic disease that its roots and progression patterns are not fully understood. Predicting the progression of T2D can inform timely and more effective interventions to prevent or manage the disease. In this study, we have used a dataset related to 63 patients with T2D that includes the data from two different types of wearable devices worn by the patients: continuous glucose monitoring (CGM) devices and activity trackers (ActiGraphs). Using this dataset, we created a model for predicting the levels of four major biomarkers related to T2D after a one-year period. We developed a wide and deep neural network and used the data from the demographic information, lab tests, and wearable sensors to create the model. The deep part of our method was developed based on the long short-term memory (LSTM) structure to process the time-series dataset collected by the wearables. In predicting the patterns of the four biomarkers, we have obtained a root mean square error of ±1.67% for HBA1c, ±6.22 mg/dl for HDL cholesterol, ±10.46 mg/dl for LDL cholesterol, and ±18.38 mg/dl for Triglyceride. Compared to existing models for studying T2D, our model offers a more comprehensive tool for combining a large variety of factors that contribute to the disease.

•Health informatics

## 1 Introduction

Diabetes mellitus refers to a group of chronic metabolic diseases resulting from problems in the production and processing of insulin production and functionality. Among various types of diabetes, the most common is type 2 (T2D), related to the condition that the cells are unable to respond adequately to normal levels of insulin (insulin resistance). In 2014, 8.5% of worldwide adults were diagnosed with some type of diabetes. With close to 90% of the diabetic patients diagnosed with T2D, the number of individuals with T2D has reached close to 380 million in 2014 and expected to exceed 450 million in 2030 [1]. In diabetic patients, the blood glucose levels (a critical indication of the body's normal condition) can exceed from its safe level. Over time, this can result in serious damages to major organs such as eyes and diseases such as cardiovascular disease. Lifestyle has been shown to significantly affect the risk of T2D and its progression patterns [2]. T2D requires constant management and care, and the progressive nature of the T2D makes this management and care very challenging. This constant care includes diet management, physical exercises, drug therapy, and insulin delivery.

Various factors are believed to determine the progression patterns of T2D. The short-term blood sugar concentration is one of the most important indexes for the providers to estimate the progression of the disease [3]. The blood glucose level can function as a momentary indicator for the overall state of a diabetic patient, and Glycated hemoglobin (HbA1c) is generally being used among medical researchers as a criterion for long term progression of the disease [4]. Considering the importance of effective control and management of T2D, modeling the progression of this disease can have a great impact on improving the effectiveness of interventions on patients with diabetes specifically if applied early enough. This type of timely intervention can be achieved through early identification of the changes in the pattern of the disease and possibly lead to decreasing the progression pace of the disease, and consequently expanding the life expectancy of patients.

Machine learning (ML) techniques offer great tools for modeling the progression of T2D. Among various forms of ML techniques that have been used for studying chronic diseases [5], the family of deep learning models is especially well suited for modeling the progression of T2D. This is mainly due to their ability to effectively identify the complex patterns among a large set of risk factors features. The complex nature of T2D is partly because of the various factors that contribute to the disease. To create models that can effectively capture the complete picture of T2D progression, it is crucial to have access to the datasets that contain many representatives of these various factors. Wearable sensors offer a great tool to capture rich datasets containing continuous (longitudinal) physiological signals.

In this study, we have used a dataset from two types of wearable sensors. The first one was Continuous Glucose Monitoring (CGM) that has revolutionized the process of monitoring blood glucose in recent years. CGM is a wearable non-invasive device that measures the glucose concentration of the blood in 1 to 5 minutes intervals, and therefore can effectively capture the fluctuations in blood glucose levels [6]. The second source of wearable sensor data in this project was a set of activity monitoring sensors. Generally, these wearable sensors are placed on the wrist or hip of an individual and monitor physical his/her movements in three-dimensional space. While these wearable sensors collect rich information from patients, wearing them for a long period can be unpleasant and impractical. As daily activities result from routines that are imposed by fixed environmental factors such as group paternal behaviors and individual habits [7], it is expected to observe some degree of consistency in time, duration, and types of activities of a person over a year. This consistency can help the progression prediction process to rely on the smaller wearable monitored dataset and reduce the length of the required monitoring process.

While various attempts have been made to model the progression of T2B, we are not aware of any study that has used longitudinal wearable sensors data in combination with other types of discretely sampled datasets (such as lab tests) to model T2D. Specifically, the main contributions of our study are:

1) We have developed a model to predict the progression of T2D using demographic, lab tests, and wearable sensor data.

2) By evaluating our model on data collected from real T2D patients, we have shown that by monitoring the wearable CGM and activity sensed signals over a short period of time as time-series signals, we can predict patients' diabetic status in one year.

## 2 Related Works

Applying machine learning (ML) techniques to diabetes problems has been an active research area in recent years. While these studies differ in methodology and application, they all to some degree try to model the progression of T2D in various stages of the disease development. HbA1c is a widely used standard measure for estimating the progression of T2D and determining the effectiveness of a treatment, and some studies tried to combine this with other biomarkers that can predict T2D patterns. For instance, Bagherzadeh et al. [8] have provided a new method for identifying an optimal set of predictors of T2D using the wrapper method as a feature selection algorithm. In another study, George et al. [9] used a variant of Random Forests method to estimate the accuracy difference resulted by selecting various features to predict blood sugar concentration in T1D patients. Simon et al. [10] used unsupervised association rules to derive the connections between diabetes risk factors.

ML-based methods have been also used to detect the likelihood of diabetes incidence for a non-patient or model the disease progress of the disease for diagnosed patients over a certain period. This problem has been mostly considered as a classification problem in the literature. In a survey study, Kavakiotis et al. [5] have compared different prediction methods in five diabetes studies. These methods were Logistic Regression, Naive Bayes, Linear Discriminant Analysis, Random Forest, Artificial Neural Networks, and Support Vector Machines. They showed that by selecting a similar set of features, SVM leads to the best prediction performance. In another study, Casanova et al. [11] presented a method based on Random Forests and demonstrated the higher robustness of their method against overfitting versus other comparable techniques in diabetes prediction. Specifically, their classification with Random Forests outperformed Logistic Regression in exploiting all the 93 features that were collected from 3600 African-American patients over a period of 8 years. Ensemble techniques to combine multiple learning methods to improve the classification performance has been also used in diabetes research. Anderson et al. [12] used a Bayesian scoring algorithm to model the space of events for a diabetes patient. Ozcift et al. [13] used Rotation Forests as the ensemble method to combine 30 ML algorithms for medical diagnosis on several diseases including diabetes.

Additionally, identifying the short- and long-term complications of Diabetes has been the topic of several studies. Many studies have focused on modeling the complex relations between diabetes and other conditions such as high blood pressure and obesity to chronic diseases such as Alzheimer's, cancer, obstructive sleep apnea, and eye damage [5]. For instance, Sudharsan et al. [14] used SVM and random forests to predict the occurrence of hypoglycemia (low blood sugar) among T2D patients, and San et al. [15] utilized heart rate signals using a deep learning framework to model the occurrence of hypoglycemia in children diagnosed with T1D. Ibrahim et al. [16] combined fuzzy logic classifiers and a clustering method to classify diabetic eye damage.

Among various types of ML methods used in the diabetes research domain, deep learning-based methods have received a lot of attention. In studying chronic conditions like diabetes, processing the sequential time-series data can be done using Recurrent Neural Networks (RNNs). RNNs are a family of deep neural networks that can extract dynamic temporal patterns from a given data sequence. For instance, for diagnosis applications, Lipton et al. [17] were one of the first to show the higher accuracy of LSTM-RNN for a classification task of time-series clinical/healthcare data by comparing the performance of the model with a multi-layer perceptron hand-engineered classifier. Choi et al. [7] used the clinical visit records of the patients in an RNN (Gated Recurrent Units) model to predict their most likely diagnoses and treatment. Their method showed a higher performance compared to Logistic Regression and multi-layer perceptron on the same dataset. Swapna et al. [18] used Heart Rate Variability (HRV) data analysis to diagnose diabetes using deep learning techniques. They achieved an accuracy of 95% using a 5-LSTM CNN with SVM. Specifically, this type of methods is still new to the area of T2D research.

## 3 Dataset

The dataset that was used in this study is collected by the co-authors at the University of Delaware and in collaboration with the Christiana Care Health System in DE, USA. All procedures for accessing and analyzing the data from this study was approved by the University of Delaware Institutional Review Board. The data was obtained from a group of patients diagnosed with T2D as part of a year-long study. The main study focused on psychological, behavioral, and interpersonal patterns among T2D patients and their spouses, and the role of collaborations between the couples coping with the disease complications. The T2D patients were in 33-78 years old age range. The records of the patients' non-diabetic spouses were not used in this current work. For each patient, various types of data were collected, including a baseline assessment involving clinical questionnaires and blood and urine tests. After seven days, the daily diary data were collected from the wearable devices that were given to the participants at the time of finishing the baseline measurements. A second blood and urine sample were collected one year after the baseline assessment. Three major components of this dataset are presented in the following.

### 3.1 CGM Measurements

In the baseline assessment visit of the study, each patient was equipped with a Dexcom G4 Platinum Professional CGM device. Each patient was trained on operating the device (e.g., how to wear the sensors and the proper care to obtain reliable readings). The CGM monitored the glucose level of each subject for seven consecutive days and recorded the glucose levels in five-minute intervals. The collected data by the end of the seven-day period contained between 1445 to 2016 consecutive CGM measurements for each patient. Out of the 63 patients in our dataset, 9 did not have any record for their second HbA1c measurement, and therefore could not be used in evaluating the performance of our models. Moreover, 4 participants had a significantly smaller number of CGM measurements compared to others. Since the length of the time-series input needed to be fixed in our model, these small sequences would enforce a big portion of other sequences to be cut. Therefore, we decided to remove these 4 records as well. As a result, 50 sequences of CGM remained for evaluation. The CGM series of data were all shortened to the size of the smallest one which was 1445. Therefore, the dimensions of the CGM data were [50×1445×1].

### 3.2 ActiGraph Data

During the same seven consecutive days of collecting the CGM measurements, body movements and physical activity patterns of the patients were recorded as well. Each patient wore an ActiGraph device on his/her non-dominant wrist. ActiGraph recorded the physical movements of each subject in three dimensions of the space (x, y, and z). We have used the ActilLife software (the commercial software provided by the manufacture of ActiGraph) to initially process the raw data into interpretable information for each patient. To include enough information about activity signal patterns, we have chosen 30 seconds as the activity processing epoch. This means that the ActiLife software divided the entire duration of data recording into 30 second intervals and extracted the physical activity information on each of those intervals. In the measured vector, the first three values ($D_x, D_y, D_z$) were the aggregated physical movements in each of the three coordinates in the corresponding interval. The fourth value ($S$) was the calculated traveled steps. The next three values ($I_{sit}, I_{std}, I_{lie}$) were inclinometer indicators, where sitting, standing, and lying activity times are saved in their corresponding fields. The eighth field ($I_{off}$) shows the undetected activity moments of a subject at a given interval. Therefore, the activity data of a subject can be presented by an eight-field vector for every 30 seconds.

### 3.3 Demographic and Lab Test Data

In addition to the above data, a rich set of demographic and clinical information was collected from each subject twice over the course of the study. At baseline (beginning of the one-year study), body weight, height, and waist circumference were measured for each participant. Participants went through lab blood tests and their biomedical indexes were recorded. These biomedical indexes included HbA1c, cholesterol, triglycerides, HDL cholesterol, LDL cholesterol, non-LDL cholesterol, and VLDL cholesterol. Also, urine samples were examined to determine creatinine and microalbumin levels. One year after the baseline assessment, these biomedical indexes were remeasured. As a criterion for the degree of T2D severity, HbA1c is an accepted standard, and the improvement/deterioration tracing of the disease could be determined using the HbA1c levels.

## 4 Methodology

The proposed framework follows a *wide and deep* structure to utilize the inherently different types of data that we had in this dataset with the goal of boosting the prediction performance. Deep neural networks (here in the form of RNN) are useful in learning the patterns from time-dependent values of the signals.

The *wide* and *deep* model [19] co-trains a deep learning network with a linear model. This method was selected for this study as we wanted to combine demographic data and lab tests with continuous time-series information. The deep part relies on time-varying value changes patterns of CGM and ActiGraph, and the wide part considers the other measured data for each patient. The idea was to effectively balance the information from each part, and use demographic data and lab test results of the patients, coupled with the features that have been extracted from the time-series wearable data to predict the HbA1c value change. Figure 1 shows an overview of the proposed predictive framework for HbA1c estimation. In this framework, the activity data was first transformed into a sequence of moving averages and then it was merged with the CGM sequence. In the following sections, we explain the procedure that was used in each of the components of this wide and deep framework.

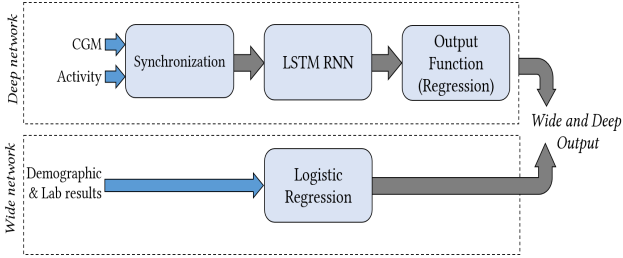

**Figure 1: The Wide and Deep learning network used in this study.**

### 4.1 Signal Synchronization

As shown in Figure 1, two signals of wearable measurements (CGM and ActiGraph) must be synchronized prior to using them in the learning network. Both CGM and ActiGraph data collection took place during the same seven-day period. There are two challenges regarding the measurement of time points in these two value sequences. First, the exact starting point of data measurements for the two signals did not match. Second, the CGM recordings were collected much more frequently compared to the activity ones. To be able to have a consistent time-series signal that carries meaningful patterns of patients' physiological information, these signals should have been first synchronized. Assuming the measurement time points ($t_i$), the CGM signal ($S_C$) may consist of $n$ (1445<n<2016) number of $C_{t_i}$ measurements, where the number of measurements varied for different patients can be shown as: $S_C = (C_{t_1}, ... C_{t_n})$. Similarly, for the ActiGraph signal ($S_A$), $m$ (15653<m<20160) number of $A_{\theta_i}$ measurements were available, and can be shown as:

$$S_A = (A_{\theta_1}, ... A_{\theta_m})$$

Since the number of measured values for both CGM and ActiGraph was not fixed among the patients, each patient had a different pair of $m$ and $n$. The following six steps were taken to generate synchronized CGM/ActiGraph signals for each patient by utilizing a *moving average with overlapping windows* method. (1) All of the CGM measurement time points $t_i$s, without any recorded ActiGraph measurement before and after them were removed. This resulted in excluding the CGM measurements that could not have an activity signal alongside them. (2) From the remaining $t_i$s, the earliest time point was selected (*i=1*). (3) The closest activity-measured time point was then detected (shown as $\theta_j$ in Figure 2). (4) The $Window_i$ of |W| activity measurements was formed. This window contained |W| closest activity time points to $t_i$. This way the average activity vector $\overline{(A_{t_i})}$ was calculated as:

$$\overline{A_{t_i}} = Average(A_{\theta_j}), where\ \theta_j \in Window_i$$

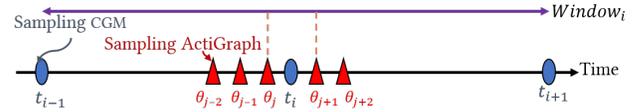

**Figure 2: Sensor measurements time points, and the window for averaging the activity measurements.**

(5) Having the average activity vector $\overline{(A_{t_i})}$, the combined synchronized CGM/ActiGraph value in time point $t_i$ was obtained in a format of: $CA_{t_i} = C_{t_i}, \bar{A}_{t_i}$. (6) To derive the next value for the merged time-series signal, the succeeding measured time point was found *(i=i+1)*, and the above procedure was iteratively continued starting from the Step (2).

In this study, we selected 50% as the overlap ratio (the portion of a window that is included in next consecutive measurement) to balance the time-series data dependency with the information loss of averaging the activity data. Considering the timing distance of the two measurements (5 minutes for CGM, and 30 seconds for ActiGraph), the window size |W| could be calculated as:

$$|W| = (\frac{|t_{i+1} - t_i|/|\theta_{j+1} - \theta_j|}{overlap\ ratio}) = \frac{(\frac{5mins}{30secs})}{50\%} = 20$$

Having the merged and synchronized time-series wearable signal, we then used them in our deep learning network.

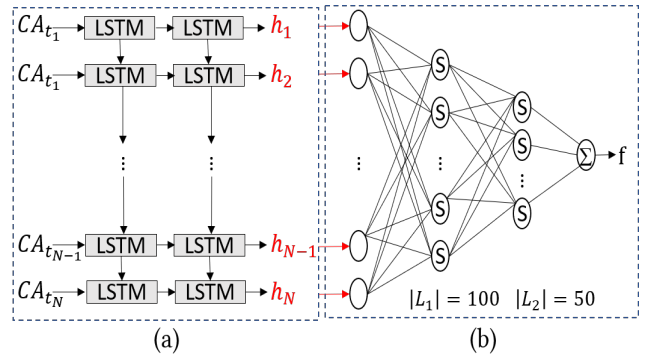

**Figure 3: (a) The structure of the LSTM RNN used for learning the patterns of the wearable data, (b) The architecture of the auxiliary network used between the RNN and output layers of our model.**

### 4.2 Long Short-Term Memory Network

Long short-term memory (LSTM) units are a specific type of RNNs that can analyze and classify a sequence of temporal data signals of any length. An LSTM unit consists of input, output, memory, and forget gates. The memory gates of an LSTM can remember the values over time. The other three gates are similar to general neuron nodes with weighted connections and an activation function. Because of their memorization capability, LSTM networks can be used to process and learn time-series signals where the sequential data points are correlated.

The proposed architecture of the RNN model consists of one input layer, two hidden layers of LSTM blocks, and one output layer (Figure 3.(a)). The input layer consists of $N$ LSTM units for $N$ time points in the time-series CGM/ActiGraph signal. Here, the input gate of the $i$th LSTM unit will be connected to the signal value at the time point $t_i$ ($CA_{t_i}$), which is a row vector with 9 features as the input of each LSTM unit. We can formulate the prediction function as:

$$\hat{\Delta}HBA1C = f(CA_{t_1}, CA_{t_2}, \ldots, CA_{t_N})$$

where the output $\hat{\Delta}HBA1C$ is the estimated change of a patient's HbA1c after one year, and the function $f$ represents the proposed model.

The desired output of the function $f$ is a single value of $\hat{\Delta}HBA1C$. However, the output of the RNN network is a time-series signal. This time-series signal should be converted to the format of a logistic regression function to obtain the desired output format. Therefore, an auxiliary network was used to process the outputs of the LSTM network (Figure 3.(b)). In this architecture, $N$ time-series outputs of the LSTM network were connected to 100 sigmoid activation blocks ($S$) of the first hidden layer of the output architecture. These 100 blocks were then connected to 50 other sigmoid activation blocks ($S$) in the second hidden layer. Finally, these blocks were linked to a linear block ($\Sigma$) to generate the output of the deep part.

### 4.3 Learning from Demographic Data and Lab Test Results

The demographic data and lab test results of the patients were then added to our deep model. The related features included: 1) height, 2) weight (kg), 3) age (years), 4) waist circumference (m), 5) Triglycerides (TC, mg/dL), 6) LDL cholesterol (mg/dL), 7) HDL cholesterol (mg/dL), and 8) VLDL cholesterol (mg/dL) of the patients. As these values for features were recorded at the baseline of the experiment, each feature had a single value for each patient and did not have any time-dependent component. To learn from this data, a logistic regression model was used forming the *wide* part of our proposed model.

## 5 Model Evaluation

All of the experiments were run on the Tensorflow platform [20] with Keras neural-network library [21]. To combine the wide network with the deep part, the *Linear Combined Classifier* in Tensorflow was used. This library offers a general framwrok for implementing wide and deep structures. To compare the performance of different parts of our model, we have performed a set of experiments using 1) only the demographics and lab test data, 2) only the CGM data, CGM and ActiGraph data, and 3) using the full dataset containing the wearable, demographics and lab test data. The models were evaluated in both regression and classification settings. The regression task refers to predicting the value of HbA1c difference, and the classification task refers to predicting whether the HbA1c value was increased or decreased. In the case of classification, the two classes of changes in HbA1c value were assumed as: 1) Improvement: ΔHbA1c≤ 0, and 2) Deterioration: ΔHbA1c>0. Besides HbA1c, which is the primary indicator of the state of T2D, we have also used our model to predict several other key biomarkers that are generally considered important in T2D control.

**Estimating HbA1c using Demographic and Lab Test Results** – In this step, the wearable sensor measurements were disregarded, and the prediction was made based on the demographic data and blood lab test results. For measuring the accuracy of the regression tasks (predicting the difference in HbA1c), we used the Root Mean Squared Error (RMSE) defined as:

$$RMSE = \mathrm{sqrt}\left(\frac{\sum_{i=1}^{N}\left(\hat{\Delta}HBA1C - \Delta HBA1C\right)^2}{N}\right)$$

where $\Delta HBA1C$ is the measured change in patients HbA1c, $\hat{\Delta}HBA1C$ is the predicted change in HbA1c, and N is the number of predicted records. The size of the input data in this experiment was [50×8] as we had 50 patients and 8 measurements for each. The prediction led to an RMSE of 2.444 with a standard deviation of σ=0.302. Since RMSE is a range dependent metric, the value ranges in the dataset are also provided to illustrate the quality of each resulted RMSE. For each biometric parameter such as HbA1c, the value range is equal to the difference between the maximum and minimum monitored values between all the patients. For the binary classification, an accuracy of 74.87% was obtained.

**Estimating HbA1c using CGM Only** – In this experiment, the RNN network was trained using the CGM signals. As previously described, the dimensions of the CGM data were [50×1445×1], as we had 1445 CGM recordings for each patient. For this experiment, we used 50 epochs for training and 5-fold cross validation, and obtained an average of RMSE=1.772 (σ=0.138). Additionally, the accuracy achieved for the classification task was equal to 81.23%. In other words, for 81.23% of the diagnosed patients, the proposed model was able to predict the improvement or deterioration status of a patient after one year correctly.

**Estimating HbA1c using CGM and ActiGraph** – In the third experiment, the RNN model was trained using the combined synchronized dataset of CGM and ActiGraph values. Data from 50 patients were used in the training and test process with 5-fold cross-validation. Having the CGM and ActiGraph signals concatenated to each other, the dimension of input data was [50×1445×9]. Again, 50 epochs were used for the experiment. The result showed an average of RMSE=1.673 (σ=0.126). Also, we have obtained a classification accuracy of 86.04% in predicting the improvement or deterioration classes.

**Estimating HbA1c using the Wide and Deep model** – The goal of this phase of the study was to evaluate the prediction performance of the wide and deep model that used both the time series and shallow demographic and lab test results of the patients. The size of input data for the wide part was [50×1445×9] and the size for the shallow part was [50×8]. We have obtained an RMSE of 1.668 (σ=0.123), which presents a slightly improved prediction performance compared to applying the wide only part of the data. For the improvement or deterioration classification task an accuracy of 86.04% was achieved.

**Estimating other biomedical values** – In the last experiment, we used the proposed model to predict HDL Cholesterol, LDL Cholesterol, and Triglyceride changes with similar experimental setup as above. The RMSE of 6.220, 10.458, and 18.377 were obtained for HDL, LDL, and Triglycerides respectively. Table 1 shows a summary of the performance of the proposed method and the obtained results in different experiments.

| Index | Signal | Size | RMSE | RMSE Range | Accuracy |
|---|---|---|---|---|---|
| HBA1c | C | [50×1445×1] | 1.722 | 0.170 | 81.23% |
| HBA1c | C, A | [50×1445×9] | 1.673 | 0.166 | 86.04% |
| HBA1c | D, L | [50×8] | 2.444 | 0.242 | 74.87% |
| HBA1c | C, A, D, L | [50×1445×9] +[50×8] | 1.668 | 0.165 | 86.04% |
| HDL | C, A | [59×1445×9] | 6.220 | 0.094 | 80.45% |
| LDL | C, A | [59×1445×9] | 10.458 | 0.070 | 81.39% |
| TC | C, A | [59×1445×9] | 18.377 | 0.029 | 89.21% |

**Table 1: The normalized error rates and prediction accuracy of predicted indexes. Index, signal type (C: CGM, A: Activity, D: Demographic, L: Lab results), size of input data, RMSE are shown. Normalized error rates (RMSE divided by Range), and improvement vs. deterioration classification accuracy are also shown.**

## 6 Conclusion

In this study, we presented a novel method for predicting the progression patterns of T2D by combining the wearable sensors data with demographic and lab test data of the patients. CGM and activity records of a group of T2D patients in a 7 days period were synchronized and used as input time-series signals for training the deep part of the model (an LSTM RNN network). In estimating the HbA1c within one year, an RMSE of 1.668 was obtained. We could achieve significantly higher accuracy than using only the demographic and lab results, which are commonly used in similar studies of diabetes.

For the binary classification, which can have a significant clinical value in the prognosis of the patients' disease, the accuracies of 86.04% and 89.21% respectively for HbA1c and Triglycerides show the potential of the model in separating the improving and deteriorating patients. Using only the CGM signals, an RMSE of 1.722 was achieved. Moreover, an RMSE of 6.220 for HDL cholesterol, 10.458 for LDL cholesterol, and 18.377 for Triglyceride were achieved. These results show that the proposed model can effectively utilize various types of features collected from a patient and achieve an acceptable error rate.

Since our model relies on balancing the learning information from two class of intrinsically independent data types, we expect that higher prediction performance will be achievable by fusing our current dataset with larger relevant datasets.

**ACKNOWLEDGMENTS**
This work was partly funded by the US NIH/NIDDK grant, R21DK098679.

**REFERENCES**
[1] W.H.O World Health Organization (2018).
[2] Wu, Y., Ding, Y., Tanaka, Y. and Zhang, W. Risk factors contributing to type 2 diabetes and recent advances in the treatment and prevention. *International journal of medical sciences*, 11, 11 (2014), 1185-1200.
[3] Sartore, G., Chilelli, N. C., Burlina, S. and Lapolla, A. Association between glucose variability as assessed by continuous glucose monitoring (CGM) and diabetic retinopathy in type 1 and type 2 diabetes. *Acta Diabetologica*, 50, 3 (June 01 2013), 437-442.
[4] The Relationship of Glycemic Exposure (HbA1c) to the Risk of Development and Progression of Retinopathy in the Diabetes Control and Complications Trial. *Diabetes*, 44, 8 (1995), 968-983.
[5] Kavakiotis, I., Tsave, O., Salifoglou, A., Maglaveras, N., Vlahavas, I. and Chouvarda, I. Machine Learning and Data Mining Methods in Diabetes Research. *Computational and Structural Biotechnology Journal*, 15 (01/ 2017).
[6] Desalvo, D. and Buckingham, B. Continuous Glucose Monitoring: Current Use and Future Directions. *Current diabetes reports*, 13 (08/ 2013).
[7] Gardner, B., Lally, P. and Wardle, J. Making health habitual: The psychology of 'habit-formation' and general practice. *The British journal of general practice : the journal of the Royal College of General Practitioners*, 62 (12/ 2012), 664-666.
[8] Bagherzadeh Khiabani, F., Ramezankhani, A., Azizi, F., Hadaegh, F., Steyerberg, E. and Khalili, D. A tutorial on variable selection for clinical prediction models: Feature selection methods in data-mining could improve the results. *Journal of clinical epidemiology* (10/ 2015).
[9] Georga, E. I., Protopappas, V. C., Polyzos, D. and Fotiadis, D. I. Evaluation of short-term predictors of glucose concentration in type 1 diabetes combining feature ranking with regression models. *Medical & Biological Engineering & Computing*, 53, 12 (Dec/01 2015), 1305-1318.
[10] J Simon, G., Schrom, J., Castro, M., Li, P. and J Caraballo, P. Survival Association Rule Mining Towards Type 2 Diabetes Risk Assessment. *AMIA, Annual Symposium proceedings*, 2013 (11/ 2013), 1293-1302.
[11] Casanova, R., Saldana, S., L. Simpson, S., Lacy, M., R. Subauste, A., Blackshear, C., Wagenknecht, L. and G. Bertoni, A. Prediction of Incident Diabetes in the Jackson Heart Study Using High-Dimensional Machine Learning. *PLOS ONE*, 11 (10/ 2016).
[12] Anderson, A. E., Kerr, W. T., Thames, A., Li, T., Xiao, J. and Cohen, M. S. Electronic health record phenotyping improves detection and screening of type 2 diabetes in the general United States population: A cross-sectional, unselected, retrospective study. *Journal of Biomedical Informatics*, 60 (apr/ 2016), 162-168.
[13] Ozcift, A. and Gulten, A. Classifier ensemble construction with rotation forest to improve medical diagnosis performance of machine learning algorithms. *Computer methods and programs in biomedicine*, 104 (04/ 2011), 443-451.
[14] Sudharsan, B., Peeples, M. and Shomali, M. Hypoglycemia Prediction Using Machine Learning Models for Patients With Type 2 Diabetes. *Journal of diabetes science and technology*, 9 (10/ 2014).
[15] San, P., Ling, S. H. and Nguyen, H. Deep learning framework for detection of hypoglycemic episodes in children with type 1 diabetes, 2016 (2016), 3503-3506.
[16] Ibrahim, S., Chowriappa, P., Dua, S., Acharya, U. R., Noronha, K., Bhandary, S. and Mugasa, H. Classification of diabetes maculopathy images using data-adaptive neuro-fuzzy inference classifier. *Medical & Biological Engineering & Computing* (06/ 2015).
[17] Lipton, Z., Kale, D., Elkan, C. and Wetzel, R. Learning to Diagnose with LSTM Recurrent Neural Networks (11/ 2015).
[18] G, S., R, V. and K.P, S. Diabetes detection using deep learning algorithms. *ICT Express*, 4, 4 (dec/ 2018), 243-246.
[19] Cheng, H.-T., Koc, L., Harmsen, J., Shaked, T., Chandra, T., Aradhye, H., Anderson, G., Corrado, G., Chai, W., Ispir, M., Anil, R., Haque, Z., Hong, L., Jain, V., Liu, X. and Shah, H. Wide & Deep Learning for Recommender Systems. *Proceedings of the 1st Workshop on Deep Learning for Recommender Systems* (2016), 7-10.
[20] Abadi, M., Agarwal, A., Barham, P. and Brevdo, E. TensorFlow: a system for large-scale machine learning. *12th USENIX conference on Operating Systems Design and Implementation* (2016), 265-283.
[21] Chollet, F. Keras (2015).